\documentclass[10pt,twocolumn,letterpaper]{article}

\usepackage{iccv}
\usepackage{times}
\usepackage{epsfig}
\usepackage{graphicx}
\usepackage{amsmath}
\usepackage{amssymb}
\usepackage{multirow}
\usepackage{booktabs}
\usepackage{color}
\usepackage{xcolor}
\usepackage{textcomp}
\usepackage{colortbl}
\usepackage{subcaption}
\usepackage{xr}
\usepackage{dsfont}
\usepackage[font=small,labelfont=bf]{caption}
\usepackage[pagebackref=true,breaklinks=true,letterpaper=true,colorlinks,bookmarks=false]{hyperref}

\newcommand\minisection[1]{\vspace{1mm}\noindent \textbf{#1}}

\usepackage[pagebackref=true,breaklinks=true,letterpaper=true,colorlinks,bookmarks=false]{hyperref}

\iccvfinalcopy 

\def\iccvPaperID{1616} 

\ificcvfinal\fi
\begin{document}

\title{Few-shot Object Detection via Feature Reweighting}


\author{
Bingyi Kang$^1$\thanks{Equal contribution.} ,  Zhuang Liu$^{2}{}^{*}$, Xin Wang$^2$, Fisher Yu$^2$, Jiashi Feng$^1$, Trevor Darrell$^2$ \vspace{0.25ex} \\
$^1$National University of Singapore ~~$^2$University of California, Berkeley\\
}

\maketitle

\newcommand{\bs}{YOLO-joint}
\newcommand{\bbsshort}{YOLO-ft}
\newcommand{\bbs}{YOLO-ft-full}
\newcommand{\lstd}{LSTD(YOLO)}
\newcommand{\lstdfull}{LSTD(YOLO)-full}
\newcommand{\Ours}{Ours}
\newcommand{\weightmodule}{reweighting module }

\begin{abstract}
Conventional training of a deep CNN based object detector demands a large number of bounding box annotations, which may be unavailable for rare categories. In this work we develop a \emph{few-shot} object detector that can learn to detect novel objects from only a few annotated examples. Our proposed model leverages fully labeled base classes and quickly adapts to novel classes, using a meta feature learner and a reweighting module within a one-stage detection architecture. The feature learner extracts meta features that are generalizable to detect novel object classes, using training data from base classes with sufficient samples. The reweighting module transforms a few support examples from the novel classes to a global vector that indicates the importance or relevance of meta features for detecting the corresponding objects. These two modules, together with a detection prediction module, are trained end-to-end based on an episodic few-shot learning scheme and a carefully designed loss function. Through extensive experiments we demonstrate that our model outperforms well-established baselines by a large margin for few-shot object detection, on multiple datasets and settings. We also present analysis on various aspects of our proposed model, aiming to provide some inspiration for future few-shot detection works. 
\end{abstract}


\section{Introduction}
The recent success of deep convolutional neural networks (CNNs) in object detection~\cite{ren2015faster,girshick2014rich,redmon2017yolo9000,redmon2018yolov3} relies heavily on a huge amount of  training data with accurate bounding box annotations. When the labeled data are scarce, CNNs can severely overfit and fail to generalize. In contrast, humans   exhibit strong  performance in such tasks: children can learn to detect a novel object quickly from very few given examples. Such ability of learning to detect from few examples is also desired for computer vision systems, since   some object categories naturally have scarce examples or their annotations are hard to obtain,  e.g., California firetrucks,   endangered animals or  certain medical data \cite{shen2017deep}.

\begin{figure}
    \centering
    \includegraphics[width=0.95\linewidth]{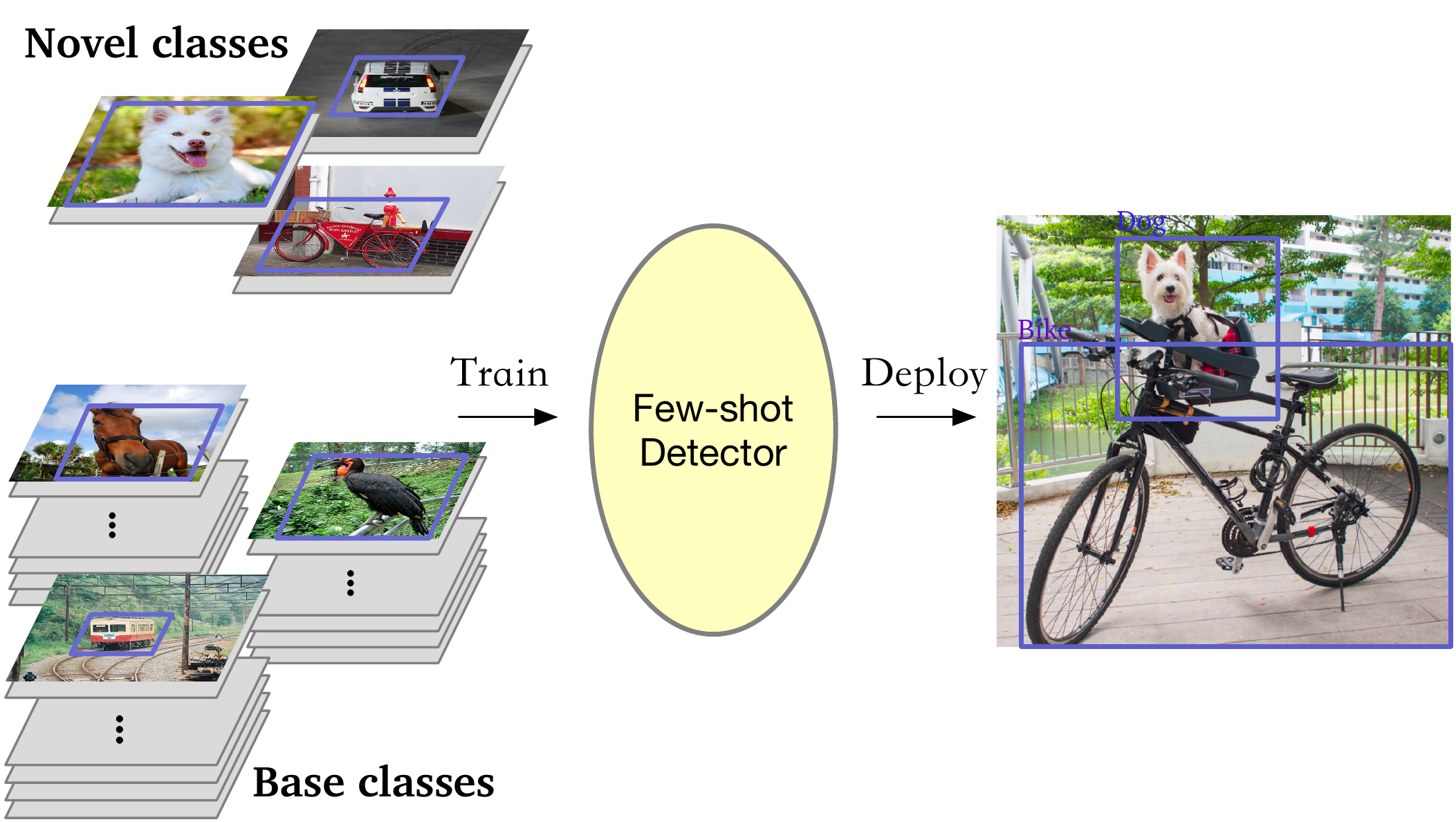}
    \caption{We aim to obtain a few-shot detection model by training on the base classes with sufficient  examples, such that the   model can learn from a few annotated examples to detect novel objects on testing images.}
    \label{fig:task}
    \vspace{-1em}
\end{figure}

 In this work, we target at the challenging \emph{few-shot object detection} problem, as shown in Fig. \ref{fig:task}. Specifically, given some base classes with sufficient examples and some novel classes with only a few samples, we aim to obtain a model that can detect both base and novel objects at test time. 
Obtaining such a few-shot detection model would be useful for many applications. Yet, effective methods are still absent. 
 Recently, meta learning \cite{vinyals2016matching, snell2017prototypical, finn2017model} offers promising solutions to a similar problem, \ie, few-shot classification. However, object detection is by nature much more difficult as it involves not only class predictions but also localization of the objects, thus off-the-shelf few-shot classification methods cannot be directly applied on the few-shot detection problem. Taking Matching Networks \cite{vinyals2016matching} and Prototypical Networks \cite{snell2017prototypical} as examples, it is unclear how to build object prototypes for matching and localization, because there may be distracting objects of irrelevant classes within the  image or no targeted objects at all. 


We propose a novel detection model that offers few-shot learning ability through fully exploiting detection training data from some base classes and quickly adapting the detection prediction network to predict novel classes according to a few support examples. The proposed model first learns meta features from base classes that are generalizable to the detection of different object classes.  Then it effectively utilizes a few support examples to identify the meta features that are important and discriminative for detecting novel classes, and adapts  accordingly to transfer detection knowledge from the base classes to the novel ones. 

\begin{figure*}[!t]
    \centering
    \includegraphics[width=\textwidth]{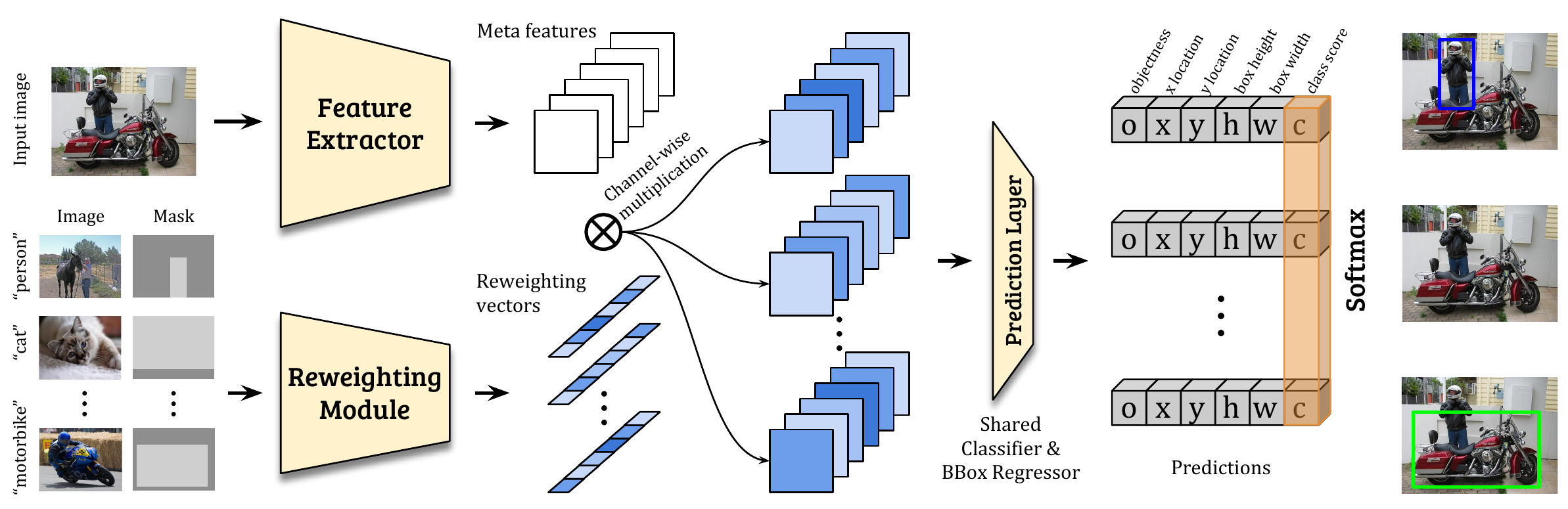}
    \caption{\textbf{The architecture of our proposed few-shot detection model. } It consists of a meta feature extractor and a reweighting module. The feature extractor follows the one-stage detector architecture and directly regresses the objectness score ($o$), bounding box location ($x,y,h,w$) and classification score ($c$). The reweighting module is trained to map support samples of $N$ classes to $N$ reweighting vectors, each responsible for modulating the meta features  to detect the objects from the corresponding class. A softmax based classification score normalization is imposed on the final output. }
    \label{fig:model}
    \vspace{-3mm}
\end{figure*}

 Our proposed model thus introduces a novel detection framework containing two  modules, \ie, a meta feature learner and a light-weight feature reweighting module. Given a query image and a few support images for novel classes, the  feature learner extracts meta features from the query image. The reweighting module learns to capture global features of  the support images  and embeds them into reweighting coefficients to modulate the query image meta features. As such, the query meta features effectively receive the support information and are adapted to be suitable for novel object detection. Then the adapted meta features are fed into a detection prediction module to predict classes and bounding boxes for novel objects in the query~(Fig.~\ref{fig:model}). 
In particular, if there are $N$ novel classes to detect, the reweighting module would take in $N$ classes of support examples and  transform them into $N$ reweighting vectors, each responsible for detecting novel objects from the corresponding class. With such class-specific reweighting vectors, some important and discriminative meta features for a novel class would be identified and contribute more to the detection decision, and the whole detection framework can learn to detect novel classes efficiently.  

The meta feature learner and the reweighting module are trained together with the detection prediction module end-to-end. To ensure few-shot generalization ability, the whole few-shot detection model is trained using an two-phase learning scheme: first learn meta features and good reweighting module from base classes; then fine-tune the detection model to adapt to novel classes. For handling difficulties in detection learning (e.g., existence of distracting objects), it introduces a carefully designed loss function. 

Our proposed few-shot detector outperforms competitive baseline methods on multiple datasets and in various settings. Besides, it also demonstrates good transferability from one dataset to another different one. Our contributions can be summarized as follows: 
\begin{itemize}
\setlength\itemsep{0em} 
    \item  We are among the first to study the problem of few-shot object detection, which is of great practical values but a less explored task than image classification in the few-shot learning literature. 
    
    \item We design a novel few-shot detection model that 1) learns generalizable meta features; and 2)  automatically reweights  the features for novel class detection by producing class-specific activating coefficients from a few support samples. 
    
    \item  We experimentally show  that our model outperforms baseline methods by a large margin, especially when the number of labels is extremely low. Our model   adapts to novel classes significantly faster.  
    
\end{itemize}


\section{Related Work}
\minisection{General object detection.}
Deep CNN based object detectors can be divided into two categories: proposal-based and proposal-free.
RCNN series~\cite{girshick2014rich,girshick2015fast,ren2015faster} detectors fall into the first category.
RCNN \cite{girshick2014rich} uses pre-trained CNNs to classify the region proposals generated by selective search~\cite{uijlings2013selective}. 
SPP-Net \cite{he2014spatial} and Fast-RCNN \cite{girshick2015fast} improve RCNN with an RoI pooling layer to extract regional features from the convolutional feature maps directly. 
Faster-RCNN \cite{ren2015faster} introduces a region-proposal-network (RPN) to improve the efficiency of generating proposals. 
In contrast, YOLO~\cite{redmon2016you} provides a proposal-free framework, which uses a single convolutional network to directly perform class and bounding box predictions. SSD~\cite{liu2016ssd} improves YOLO by using default boxes (anchors) to adjust to various object shapes. YOLOv2 \cite{redmon2017yolo9000} improves YOLO with a series of techniques, e.g., multi-scale training, new network architecture (DarkNet-19). Compared with proposal-based methods, proposal-free methods do not require a per-region classifier, thus are conceptually simpler and significantly faster. Our few-shot detector is built on the YOLOv2 architecture.

\minisection{Few-shot learning.}
Few-shot learning refers to learning from just a few training examples per class. Li~\etal~\cite{li2006one} use Bayesian inference to generalize knowledge from a pre-trained model to perform one-shot learning. Lake~\etal~\cite{lake2013one} propose a Hierarchical Bayesian one-shot learning system that exploits compositionality and causality. Luo~\etal~\cite{luo2017label} consider the problem of adapting to novel classes in a new domain. Douze~\etal~\cite{DouzeCVPR2018} assume abundant unlabeled images and adopts label propagation in a semi-supervised setting.  

An increasingly popular solution for few-shot learning is meta-learning, which can further be divided into three categories: a) Metric learning based~\cite{koch2015siamese, sung2018learning,vinyals2016matching,snell2017prototypical}. 
In particular, Matching Networks \cite{vinyals2016matching} learn the task of finding the most similar class for the target image among a small set of labeled images. Prototypical Networks \cite{snell2017prototypical} extend Matching Networks by producing a linear classifier instead of weighted nearest neighbor for each class. Relation Networks \cite{sung2018learning} learn a distance metric to compare the target image to a few labeled images. 
b) Optimization for fast adaptation. Ravi and Larochelle~\cite{ravi2016optimization} propose an LSTM meta-learner that is trained to quickly converge a learner classifier in new few-shot tasks. Model-Agnostic Meta-Learning (MAML) \cite{finn2017model} optimizes a task-agnostic network so that a few gradient updates on its parameters would lead to good performance on new few-shot tasks. c) Parameter prediction. Learnet \cite{bertinetto2016learning} dynamically learns the parameters of factorized weight layers based on a single example of each class to realize one-shot learning.

Above methods are developed to recognize novel images only, there are some other works tried to learn a model that can classify both base and novel images. Recent works by Hariharan~\etal \cite{hariharan2017low, wang2018low} introduce image hallucination techniques to augment the novel training data such that novel classes and base classes are balanced to some extend. Weight imprinting~\cite{qi2017low} sets weights for a new category using a scaled embedding of labeled examples. Dynamic-Net~\cite{gidaris2018dynamic} learns a weight generator to classification weights for a specific category given the corresponding labeled images. These previous works only tackle image classification task, while our work focuses on object detection.

\noindent\textbf{Object detection with limited labels.}
There are a number of prior works on detection focusing on settings with limited labels.
The weakly-supervised setting \cite{bilen2016weakly, diba2017weakly, song2014weakly} considers the problem of training object detectors with only image-level labels, but without bounding box annotations, which are more expensive to obtain. 
Few example object detection \cite{misra2015watch, wang2015model, dong2017few} assumes only a few labeled bounding boxes per class, but relies on abundant unlabeled images to generate trustworthy pseudo annotations for training. 
Zero-shot object detection \cite{bansal2018zero, rahman2018zero, zhu2018zero} aims to detect previously unseen object categories, thus usually requires external information such as relations between classes. Different from these settings, our few-shot detector uses very few bounding box annotations (1-10) for each novel class, without the need for unlabeled images or external knowledge.  Chen~\etal~\cite{chen2018lstd} study a similar setting but only in a transfer learning context, where the target domain images only contains novel classes without base classes.

\section{Approach}
\label{sec:method}


In this work, we define a novel and realistic setting for {few-shot object detection}, in which there are two kinds of data available for training,  \ie,  the \emph{base classes}  and the \emph{novel classes}. For the base classes, abundant annotated data are available, while only a few labeled samples are given to the novel classes \cite{hariharan2017low}. We aim to obtain a few-shot detection model that can learn to detect novel object when there are both base and novel classes in testing by leveraging knowledge from the base classes. 

This setting is worth exploring since it aligns well with a practical situation\textemdash one may expect to deploy a pre-trained detector for new classes with only a few labeled samples. More specifically, large-scale object detection datasets (e.g., PSACAL VOC, MSCOCO) are available to pre-train a detection model. However, the number of object categories therein is quite limited, especially compared to the vast object categories in real world. Thus, solving this few-shot object detection problem  is heavily desired.

\subsection{Feature Reweighting for Detection} 

Our proposed few-shot detection model introduces a meta feature learner $\mathcal{D}$ and a \weightmodule $\mathcal{M}$ into a one-stage detection framework. In this work, we adopt the proposal-free detection framework YOLOv2~\cite{redmon2017yolo9000}. It directly regresses features for each anchor to detection relevant outputs including classification score and object bounding box coordinates through a detection prediction module $\mathcal{P}$.
As shown in Fig. \ref{fig:model}, we adopt the backbone of YOLOv2 (\ie, DarkNet-19) to implement the meta feature extractor $\mathcal{D}$, and follow the same anchor setting as YOLOv2. As for the \weightmodule $\mathcal{M}$, we carefully design it to be a light-weight CNN for both enhancing efficiency and easing its learning. Its architecture details are deferred to the supplementary due to space limit.  


The meta feature learner $\mathcal{D}$ learns how to extract meta features for the input query images to detect their novel objects. The \weightmodule $\mathcal{M}$, taking the support examples as input, learns to embed support information into reweighting vectors and adjust contribution of each meta feature of the query image accordingly for following detection prediction module $\mathcal{P}$. With the \weightmodule, some meta features informative for detecting novel objects would be excited and thus assist detection prediction.

Formally, let $I$ denote an input query image. Its corresponding meta features $F\in\mathbb{R}^{w\times h\times m}$ are generated by   $\mathcal{D}$:  $F = \mathcal{D}(I)$. The produced meta feature has $m$ feature maps.
We denote the support images and their associated bounding box annotation, indicating the target class to detect, as $I_i$ and $M_i$ respectively, for class $i, i = 1,\ldots,N$.  The \weightmodule $\mathcal{M}$ takes  one support image $(I_i, M_i)$  as input and embed it into a class-specific representation $w_i \in \mathbb{R}^m$ with $w_i = \mathcal{M} (I_i, M_i)$. Such embedding captures global representation of the target object w.r.t.\ the $m$ meta features. It will be responsible for reweighting the meta features and highlighting more important and relevant ones to detect the target object from class $i$. 
More specifically, after obtaining the class-specific reweighting coefficients $w_i$, our model applies it to obtain the class-specific feature $F_i$ for novel class $i$ by:
\begin{equation}
    \label{eq:reweighting}
    F_i = F \otimes w_i, ~~~ i = 1,\ldots, N,
\end{equation}
where $\otimes$ denotes channel-wise multiplication. 
We 
implement it through 1$\times$1 depth-wise convolution.

After acquiring class-specific features $F_i$, we feed them into the {prediction module} $\mathcal{P}$ to  regress the objectness score $o$, bounding box location offsets $(x,y,h,w)$, and classification score $c_i$ for each of a set of predefined anchors:
\begin{equation}
\{o_i, x_i, y_i, h_i, w_i, c_i\} = \mathcal{P}(F_i), ~~~ i = 1,\dots,N, 
\end{equation}
where $c_i$ is one-versus-all classification score indicating the probability of the corresponding object belongs to class $i$. 


\subsection{Learning Scheme}
\label{sec:training_details}
It is not straightforward to learn a good meta feature learner  $\mathcal{D}$ and \weightmodule $\mathcal{M}$  from the base classes such that they can produce generalizable  meta features and rweighting coefficients. To ensure the model generalization performance from few examples, we develop a new two-phase learning scheme that is different from the conventional ones for detection model training.

We reorganize the training images with annotations from the  base classes into multiple few-shot detection learning tasks $\mathcal{T}_j$. Each task $\mathcal{T}_j = \mathcal{S}_j \cup \mathcal{Q}_j = \{(I^j_1,M^j_1), \ldots, (I^j_N,M^j_N)\} \cup  \{(I^q_j,M^q_j)\}$ contains a support set $S_j$ (consisting of $N$ support images each of which is from a different base class) and a query set $\mathcal{Q}_j$ (offering query images with annotations for performance evaluation). 

Let $\theta_D$, $\theta_M$ and $\theta_P$  denote the parameters of meta feature learner  $\mathcal{D}$, the \weightmodule $\mathcal{M}$ and prediction module $\mathcal{P}$ respectively. We optimize them jointly through minimizing the following loss:
\begin{equation*}
\begin{aligned}
    \min_{\theta_D,\theta_M,\theta_P} \mathcal{L} :=& \sum_{j}\mathcal{L}(\mathcal{T}_j) \\
    = &\sum_j \mathcal{L}_{\text{det}} (\mathcal{P}_{\theta_P}(\mathcal{D}_{\theta_D}(I_q^j) \otimes \mathcal{M}_{\theta_M}(\mathcal{S}_j)), M^q_j).
    \end{aligned}
\end{equation*}
Here $\mathcal{L}_{\text{det}}$ is the detection loss function and we explain its details later. The above optimization ensures the model to learn good meta features for the query and reweighting coefficients for the support.  

The overall learning  procedure consists of two phases. The first phase is the  \emph{base training} phase. 
In this phase, despite abundant labels are available for each base class, we still jointly train the feature learner, detection prediction together with the \weightmodule. This is to make them coordinate in a desired way: the model needs to learn to detect objects of interest by referring to a good reweighting vector. 
The second phase is \emph{few-shot fine-tuning}. In this phase, we train the model on both base and novel classes. As only $k$ labeled bounding boxes are available for the novel classes, to balance between samples from the base and novel classes,  we also include $k$ boxes for each base class. The training procedure is the same as the first phase, except that it takes significantly fewer iterations for the model to converge.  

In both training phases, the reweighting coefficients depend on the input pairs of (support image, bounding box)  that are randomly sampled from the available data  per iteration. After few-shot fine-tuning, we would like to obtain a detection  model that can directly perform detection without requiring any support  input.  {This is achieved by setting the reweighting vector for a target class to the average one predicted by the model after taking the $k$-shot samples as input. After this, the \weightmodule can be completely removed during inference. Therefore, our model adds negligible extra model parameters to the original detector}

\minisection{Detection loss function.} 
To train the few-shot detection model, we need to carefully choose the loss functions in particular for the class prediction branch, as the sample number is very few. Given that the predictions are made class-wisely, it seems natural to use binary cross-entropy loss, regressing 1 if the object is the target class and 0 otherwise. However, we found using this loss function gave a model prone to outputting redundant detection results (e.g., detecting a train as a bus and a car). This is due to that for a specific region of interest, only one out of $N$ classes is truly positive. However, {the binary loss  strives to produce balanced positive and negative predictions}. Non-maximum suppression could not help remove such false positives as it only operates on predictions within each class.

To resolve this issue, our proposed model adopts a softmax layer for calibrating the classification scores among different classes, and adaptively lower detection scores for the  wrong classes. Therefore, the actual classification score for the $i$-th class  is given by $\hat{c}_i =  \frac{e^{c_i}}{\sum_{j=1}^N e^{c_j}}$.
Then to better align training procedure  and few-shot detection, the    cross-entropy loss  over the calibrated scores $\hat{c}_i$ is adopted:
\begin{equation}
    \mathcal{L}_c = -\sum_{i=1}^{N} \mathds{1}(\cdot, i) \log(\hat{c}_i),
\end{equation}
where $\mathds{1}(\cdot, i)$ is an indicator function for whether current anchor box really belongs to class $i$ or not.  After introducing softmax, the summation of classification scores for a specific anchor  is equal to 1, and less probable class predictions will be suppressed. This softmax loss will be shown to be superior to binary loss in the following experiments. For bounding box and objectiveness regression, we adopt the similar loss function $ \mathcal{L}_{bbx}$ and $\mathcal{L}_{obj}$ as YOLOv2~\cite{redmon2017yolo9000} but we balance the positive and negative by not computing some loss from negatives samples for the objectiveness scores. Thus, the overall detection loss function is $\mathcal{L}_{\text{det}} = \mathcal{L}_c + \mathcal{L}_{bbx} + \mathcal{L}_{obj}$.

\minisection{Reweighting module input.} The input of the reweighting module should be the object of interest. However, in object detection task, one image may contain multiple objects from different classes. To let the \weightmodule know what the target class is, in additional to three RGB channels, we include an additional ``mask'' channel ($M_i$) that has only binary values: on the position within the bounding box of an object of interest, the value is 1, otherwise it is 0 (see left-bottom of Fig. \ref{fig:model}). If multiple target objects are present on the image, only one object is used. This additional mask channel gives the reweighting module the knowledge of what part of the image's information it should use, and what part should be considered as ``background''. Combining mask and image as input not only provides class information of the object of interest but also the location information (indicated by the mask) useful for detection. In the experiments, we also investigate other input forms.

\section{Experiments}

In this section, we evaluate our model and 
compare it with various baselines, to show our model can learn to detect  novel objects significantly faster and more accurately. We use 
YOLOv2 \cite{redmon2017yolo9000} as the base detector.  Due to space limit, we defer all the model architecture and implementation details  to the supplementary material. 
The code to reproduce the results will be released at \url{https://github.com/bingykang/Fewshot_Detection}. 

\subsection{Experimental Setup}
\minisection{Datasets.} We evaluate our model for few-shot detection on the widely-used object detection benchmarks, i.e.,  VOC 2007 \cite{everingham2010pascal}, VOC 2012 \cite{everingham2015pascal}, and MS-COCO \cite{lin2014microsoft}. 
We follow the common practice \cite{redmon2017yolo9000, ren2015faster, shen2017dsod, dai2016rfcn} and use VOC 07 test set for testing while use VOC 07 and 12 train/val sets for training. Out of its 20 object categories, 
we randomly select 5 classes as the novel ones, while keep the remaining 15 ones as the base. 
We evaluate with 3 different base/novel splits. During base training, only annotations of the base classes are given. For few-shot fine-tuning, we use a very small set of training images to ensure that each class of objects only has $k$ annotated bounding boxes, where $k$ equals 1, 2, 3, 5 and 10. Similarly, on the MS-COCO dataset,  we use 5000 images from the validation set for evaluation, and the rest images in train/val sets for training. Out of its 80 object classes,  we select 20 classes overlapped with VOC  as novel classes, and the remaining 60 classes as  the base classes. 
We also consider learning the model on the 60 base classes from COCO and applying it to detect  the 20 novel objects in PASCAL VOC. This setting features a cross-dataset learning problem that we denote as \emph{COCO to PASCAL}.  

Note the testing images may contain distracting base classes (which are not targeted classes to detect)  and some images do not contain  objects of the targeted novel class. This makes the few-shot detection further challenging. 



\begin{table*}[!ht]
\centering
\footnotesize
\setlength{\tabcolsep}{0.4em}
\begin{tabular}{l|ccccc|ccccc|ccccc}
\toprule
\multicolumn{1}{c|}{} & \multicolumn{5}{c|}{Novel Set 1} & \multicolumn{5}{c|}{Novel Set 2} & \multicolumn{5}{c}{Novel Set 3} \\ \midrule
Method / Shot         & 1     & 2     & 3    & 5    & 10   & 1     & 2     & 3    & 5    & 10   & 1     & 2     & 3    & 5    & 10   \\ \midrule
\bs{}                 & 0.0   & 0.0   & 1.8  & 1.8  & 1.8  & 0.0   & 0.1   & 0.0  & 1.8  & 0.0    & 1.8   & 1.8   & 1.8  & 3.6  & 3.9  \\ 
\bbsshort{}           & 3.2   & 6.5   & 6.4  & 7.5  & 12.3 & 8.2   & 3.8   & 3.5  & 3.5  & 7.8  & 8.1   & 7.4   & 7.6  & 9.5  & 10.5 \\ 
\bbs{}                & 6.6   & 10.7  & 12.5 & 24.8 & 38.6 & 12.5  & 4.2   & 11.6 & 16.1 & 33.9 & 13.0  & 15.9  & 15.0 & 32.2 & 38.4 \\ 
\lstd{}               & 6.9   & 9.2   & 7.4 & 12.2 & 11.6 & 9.9  & 5.4   & 3.3 & 5.7 & 19.2  & 10.9  & 7.6 & 9.5  & 15.3 & 16.9 \\ 
\lstdfull{}           & 8.2  & 11.0 & 12.4 & 29.1 & 38.5 & 11.4  & 3.8  & 5.0 & 15.7 & 31.0 & 12.6  & 8.5 & 15.0 & 27.3 & 36.3 \\ \midrule
Ours                  & \textbf{14.8}  & \textbf{15.5}  & \textbf{26.7} & \textbf{33.9} & \textbf{47.2} & \textbf{15.7}  & \textbf{15.3}  & \textbf{22.7} & \textbf{30.1} & \textbf{40.5} & \textbf{21.3}  & \textbf{25.6}  & \textbf{28.4} & \textbf{42.8} & \textbf{45.9} \\ \bottomrule
\end{tabular}
\caption{Few-shot detection performance (mAP)  on the PASCAL VOC dataset. We evaluate the performance on three different sets of novel categories. Our model consistently outperforms baseline methods. 
}
\label{tab:main}
\end{table*}
\begin{table*}
    \centering
    \footnotesize
    \begin{tabular}{cl|ccc|ccc|ccc|ccc}
    \toprule
    & & \multicolumn{6}{c|}{Average Precision} & \multicolumn{6}{c}{Average Recall}\\ \midrule
    \# Shots   &  & 0.5:0.95 & 0.5 & 0.75 & S & M & L & 1 & 10 & 100 & S & M & L \\
    \midrule
    \multirow{5}{*}{10 } 
    & \bbsshort{} & 0.4 & 1.1 & 0.1 & 0.3 & 0.7 & 0.6 & 5.8 & 8.0 & 8.0 & 0.6 & 5.1 & 15.5 \\
    & \bbs{} & 3.1 & 7.9 & 1.7 & 0.7 & 2.0 & 6.3 & 7.8 & 10.5 & 10.5 & 1.1 & 5.5 & 20 \\ 
    & \lstd{} & 0.4 & 1.1 & 0.2 & 0.2 & 0.7 & 0.6 & 5.8 & 7.9 & 7.9 & 0.6 & 5.0 & 15.3 \\ 
    & \lstdfull{} & 3.2 & 8.1 & 2.1 & \textbf{0.9} & 2.0 & 6.5 & 7.8 & 10.4 & 10.4 & 1.1 & 5.6 & 19.6 \\ 
    & Ours & \textbf{5.6} & \textbf{12.3} & \textbf{4.6} & \textbf{0.9} & \textbf{3.5} & \textbf{10.5} & \textbf{10.1} & \textbf{14.3} & \textbf{14.4} & \textbf{1.5} & \textbf{8.4} & \textbf{28.2} \\
    \midrule
    \multirow{5}{*}{30 } 
    & \bbsshort{} & 0.6 & 1.5 & 0.3 & 0.2 & 0.7 & 1.0 & 7.4 &  9.4 & 9.4  & 0.4 & 3.9 & 19.3 \\
    & \bbs{} & 7.7 & 16.7 & 6.4 & 0.4 & 3.3 & 14.4 & 11.7 & 15.3 & 15.3 & 1.0 & 7.7 & 29.2 \\ 
    & \lstd{} & 0.6 & 1.4 & 0.3 & 0.2 & 0.8 & 1.0 & 7.1 & 9.1 & 9.2 & 0.4 & 3.9 & 18.7 \\ 
    & \lstdfull{} & 6.7 & 15.8 & 5.1 & 0.4 & 2.9 & 12.3 & 10.9 & 14.3 & 14.3 & 0.9 & 7.1 & 27.0 \\ 
    & Ours & \textbf{9.1} & \textbf{19.0} & \textbf{7.6} & \textbf{0.8} & \textbf{4.9} & \textbf{16.8} & \textbf{13.2} & \textbf{17.7} & \textbf{17.8} & \textbf{1.5} & \textbf{10.4} & \textbf{33.5} \\
    \bottomrule
    \end{tabular}
    \caption{Few-shot detection performance for the novel categories on the COCO dataset. We evaluate the performance for different numbers of training shots for the novel categories.}
    \vspace{-3mm}
    \label{tab:coco}
\end{table*}
\minisection{Baselines.} 
We compare our model with five competitive baselines. Three of them are built upon the vanilla YOLOv2 detector with straightforward few-shot learning strategies. The first one is to   train  the  detector on  images from the base   and novel classes together. In this way, it can learn good features from the base classes that are applicable for detecting  novel classes.  We term this baseline as \emph{YOLO-joint}. We train this baseline model with the same total iterations as ours. 
The other two YOLO-based  baselines also use two training phases  as ours. In particular,  they train  the original YOLOv2 model with the same base training phase as ours; for the few-shot fine-tuning phase, one fine-tunes the model with the same iterations as ours, giving the \emph{YOLO-ft} baseline; and one trains  the  model to fully converge, giving \emph{YOLO-ft-full}. Comparing with these  baselines  can help understand the few-shot learning  advantage of our models brought by the proposed feature reweighting method. The last two baselines are from a recent few-shot detection method, i.e., Low-Shot Transfer Detector (LSTD)~\cite{chen2018lstd}. LSTD relies on background depression (BD)  and transfer knowledge (TK) to obtain a few-shot detection model on the novel classes. For fair comparison, we re-implement BD and TK based on YOLOV2, train it for the same iterations as ours, obtaining \emph{\lstd{}};  and train it to convergence to obtain the last baseline,  \emph{\lstdfull{}}.

\subsection{Comparison with Baselines}

\minisection{PASCAL VOC.} 
We present our main results on novel classes in Table \ref{tab:main}. First we note that our model significantly outperforms the baselines, especially when the labels are extremely scarce (1-3 shot). The improvements are also consistent for different base/novel class splits and number of shots. In contrast, \lstd{} can boost performance in some cases, but might harm the detection in other cases. Take 5-shot detection as an example, \lstdfull{} brings 4.3 mAP improvement compared to \bbs{} on novel set 1, but it is worse than \bbs{} by 5.1 mAP on novel set 2.
Second, we note that \bbsshort{}/\bbs{} also performs significantly better than \bs{}. This demonstrates the necessity of the two training phases employed in our model: it is better to first train a good knowledge representation on base classes and then fine-tune with few-shot data, otherwise joint training with let the detector bias towards base classes and learn nearly nothing about novel classes. More detailed results about each class is available at supplementary material. 



\minisection{COCO.}
The results for COCO dataset is shown in Table \ref{tab:coco}. We evaluate for $k=10$ and $k=30$ shots per class. In both cases, our model outperforms all the  baselines. In particular, when the YOLO baseline is trained with same iterations with our model, it achieves an AP of less than 1\%. We also observe that there is much room to improve the results obtained in the few-shot scenario. This is possibly due to the complexity and large amount of data in COCO so that few-shot detection over it is quite challenging.

\minisection{COCO to PASCAL.} We evaluate our model using 10-shot image per class from PASCAL. The mAP of \bbsshort{}, \bbs{}, \lstd{}, \lstdfull{} are 11.24\%, 28.29\%, 10.99\% 28.95\% respectively, while our method achieves 32.29\%. The performance on PASCAL novel classes is worse than that when we use base classes in PASCAL  dataset (which has mAP around 40\%). This might be explained by the different numbers of novel classes, \ie, 20 v.s.\ 5. 

\subsection{Performance Analysis}

\begin{figure}[!ht]
    \centering
    \includegraphics[ width=0.9\linewidth]{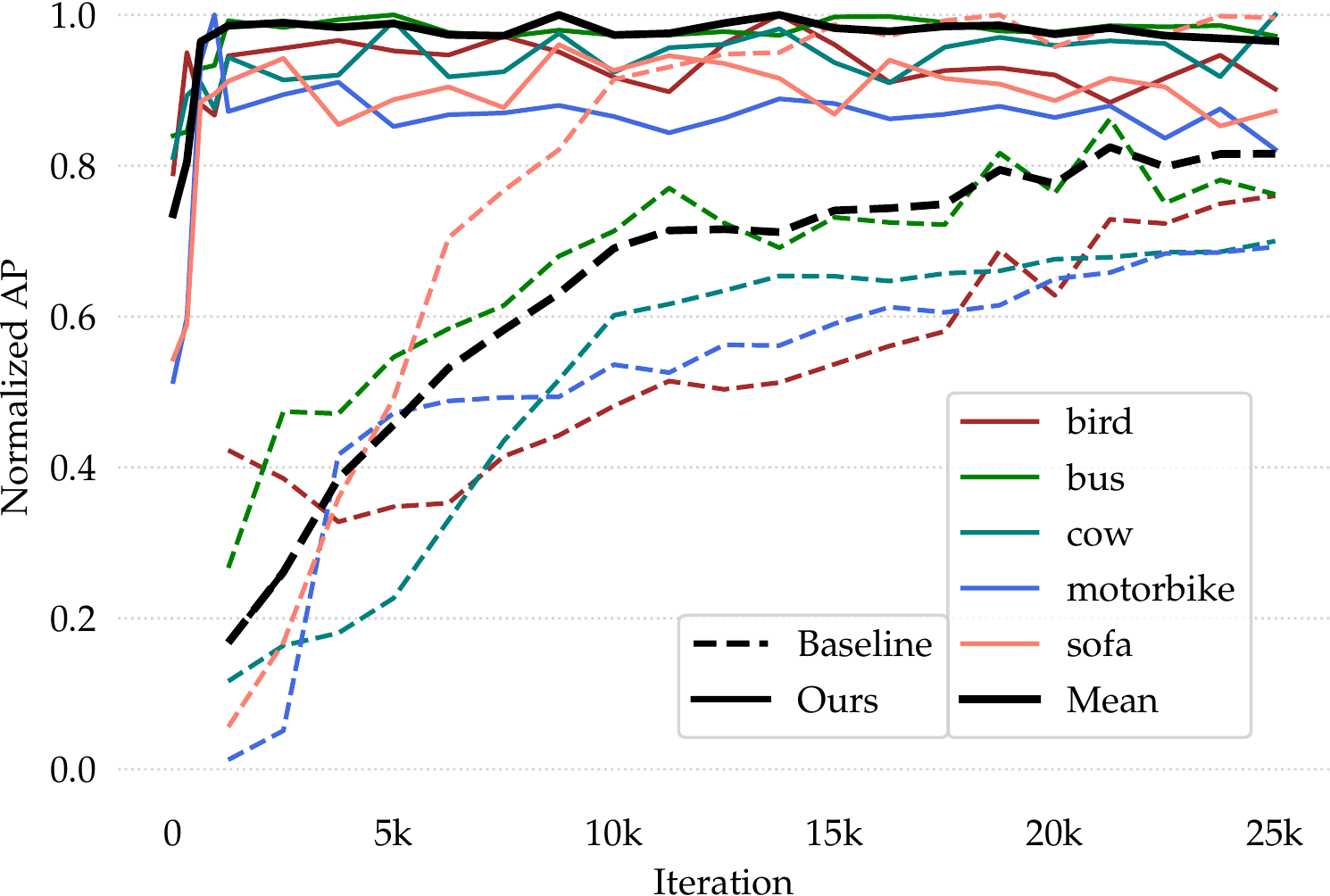}
    \caption{Learning speed comparison between our proposed few-shot detection model and the YOLO-ft-full baseline. We plot the AP (normalized by the converged value) against number of training iterations. Our model shows much faster adaption speed.}
    \label{fig:speed}
    \vspace{-4mm}
\end{figure}

\minisection{Learning speed.}
Here we analyze learning speed of our models. The results  show that despite the fact that our few-shot detection model does not consider adaptation speed explicitly in the optimization process, it still exhibits surprisingly fast adaptation ability. Note that in experiments of Table \ref{tab:main},  \bbs{} and \lstdfull{} requires 25,000 iterations for it to fully converge, while our model only require 1200 iterations to converge to a higher accuracy. When the baseline YOLO-ft and LSTD(YOLO) are trained for the same iterations  as ours, their performance is far worse. In this section, we compare the full convergence behavior of \bs{}, \bbs{}  and our method in Fig. \ref{fig:speed}. The AP value are normalized by the maximum value during the training of our method and the baseline together. This experiment is conducted on PASCAL VOC base/novel split 1, with 10-shot bounding box labels on novel classes.

From Fig. \ref{fig:speed}, our method (solid lines) converges significantly faster than the baseline YOLO detector (dashed lines), for each novel class as well as on average. For the class Sofa (orange line), despite the baseline YOLO detector eventually slightly outperforms our method, it takes a great amount of training iterations to catch up with the latter. This behavior makes our model a good few-shot detector in practice, where scarcely labeled novel classes may come in any time and short adaptation time is desired to put the system in real usage fast. This also opens up our model's potential in a life-long learning setting \cite{chen2018lifelong}, where the model accumulates the knowledge learned from past and uses/adapts it for future prediction. We also observe similar convergence advantage of our model over \bbs{} and \lstdfull{}.

\vspace{2mm}

\minisection{Learned reweighting coefficients.}
\label{sec:vis}
\begin{figure*}
    \centering
    \begin{subfigure}{0.59\textwidth}
        \includegraphics[width=\textwidth]{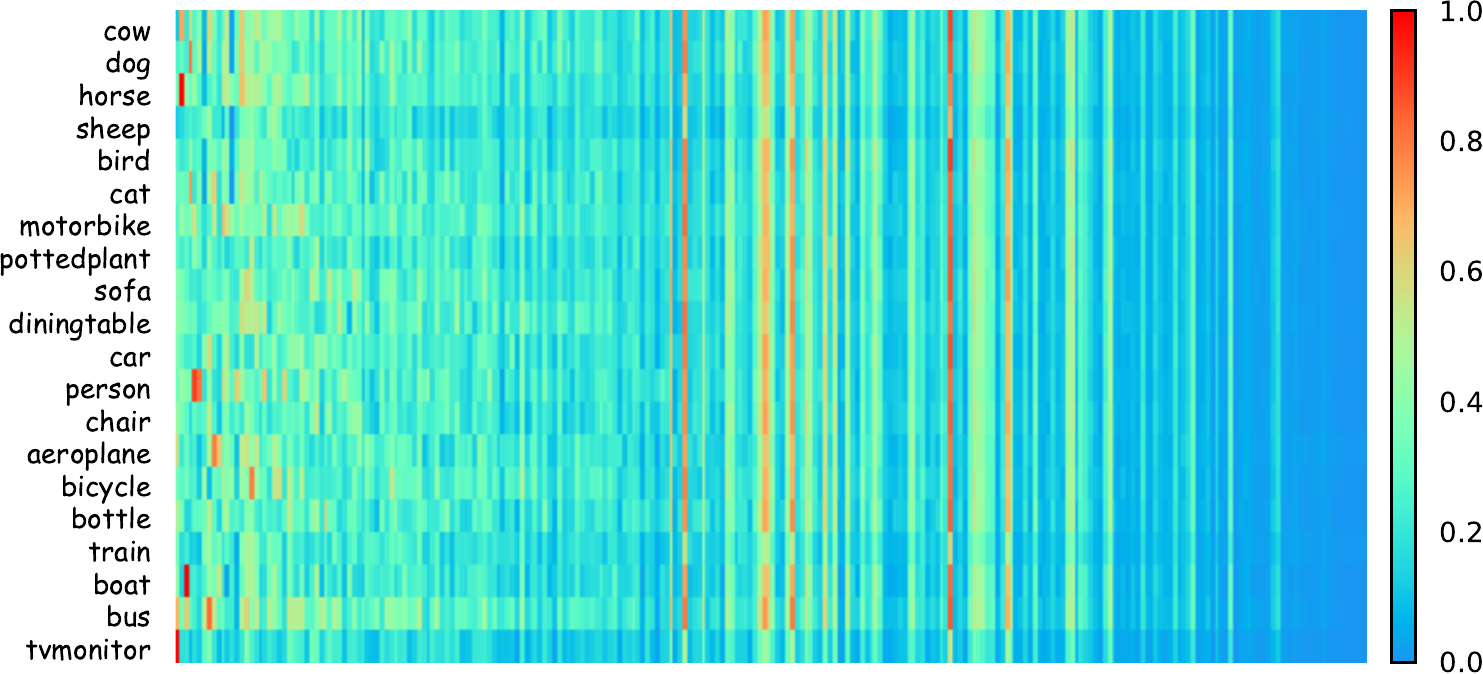}
        \caption{}
        \label{fig:heatmap}
    \end{subfigure}
    \hspace{1em}
    \begin{subfigure}{0.35\textwidth}
        \includegraphics[width=\textwidth]{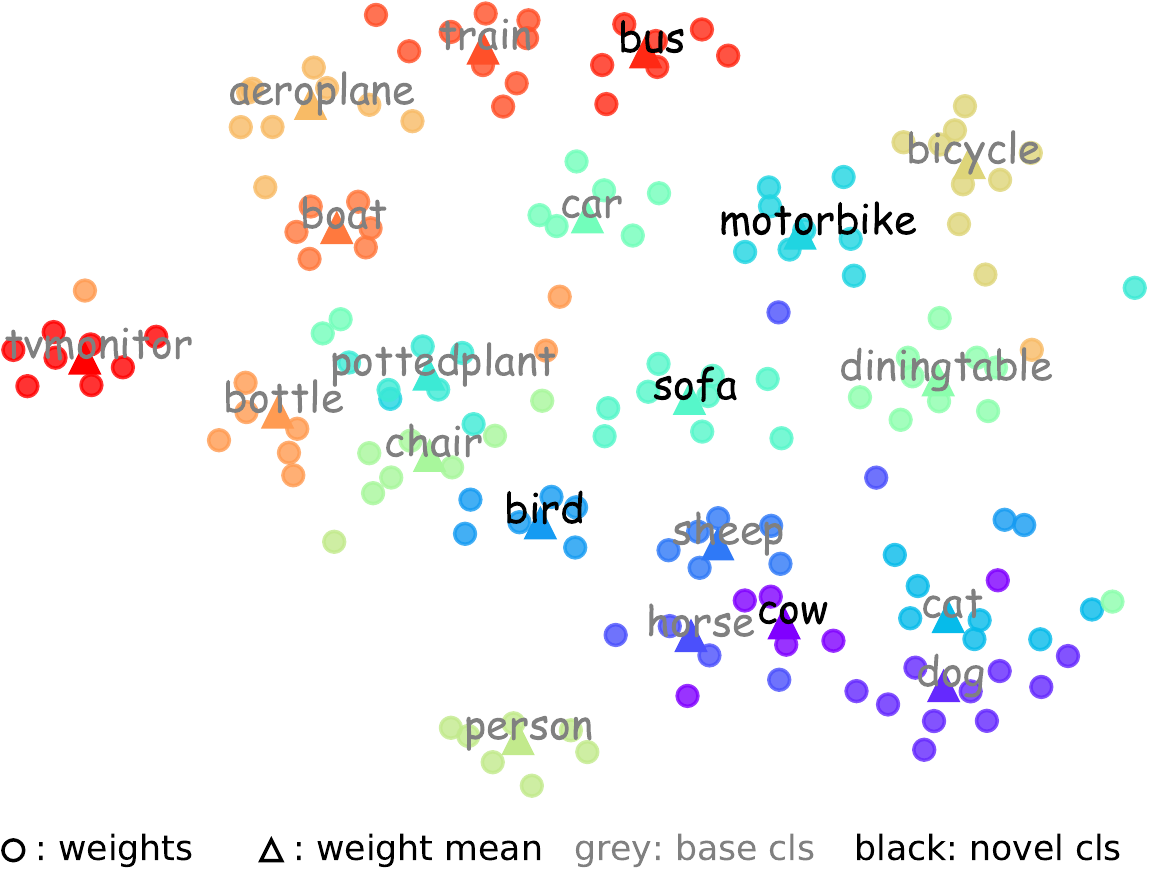}
        \caption{}
        \label{fig:tsne}
    \end{subfigure}
    \vspace{-1em}
    \caption{(a) Visualization of the reweighting coefficients (in row vectors) from the reweighting module for each class. Columns correspond to meta feature maps, ranked by variance among classes.  Due to space limit, we only plot randomly sampled 256 features. (b) t-SNE \cite{maaten2008visualizing} visualization of the reweighting coefficients. More visually similar classes tend to have closer coefficients. }
    \vspace{-4mm}
\end{figure*}
The reweighting coefficient is important for the meta-feature usage and   detection performance. To see this, we first plot the 1024-d reweighting vectors for each class in Fig. \ref{fig:heatmap}. In the figure, each row corresponds to a class and each column corresponds to a feature. The features are ranked by variance among 20 classes from left to right. We observe that roughly half of the features (columns) have notable variance among different classes (multiple colors in a column), while the other half are insensitive to classes (roughly the same color in a column). This suggests that indeed only a portion of features are used differently when detecting different classes, while the remaining ones are shared across different classes. 

We further visualize the reweighting vectors by t-SNE~\cite{maaten2008visualizing} in Fig. \ref{fig:tsne} learned from 10 shots/class on base/novel split 1. In this figure, we   plot the reweighting vector generated by each support input,  along with their average for each class. We observe that not only vectors of the same classes tend to form clusters, the ones of visually similar classes also tend to be close. For instance, the classes Cow, Horse, Sheep, Cat and Dog are all around the right-bottom corner, and they are all animals. Classes of transportation tools are at the top of the figure. Person and Bird are more visually different from the mentioned animals, but are still closer to them than the transportation tools.

\minisection{Learned meta features.}
Here we analyze the learned meta features from the base classes in the  first training stage. Ideally, a desirable few-shot detection model should preferably perform as well when data are abundant. We compare the mAP on base classes for models obtained after the first-stage base training, between our  model and the vanilla YOLO detector (used in latter two baselines). The results are shown in Table \ref{tab:base}. Despite our detector is designed for a few-shot scenario, it also has strong representation power and offers good meta features  to reach comparable performance with the original YOLOv2 detector trained on a lot of samples. 
This lays a basis for solving the few-shot object detection problem.
\begin{table}[!h]
\small
\centering
\begin{tabular}{lccc}
\toprule
              & Base Set 1 & Base Set 2 & Base Set 3 \\ \midrule
YOLO Baseline & \textbf{70.3}  & \textbf{72.2}  & 70.6  \\ \midrule
Our model     & 69.7  & 72.0  & \textbf{70.8}  \\ \bottomrule
\end{tabular}
\caption{Detection performance (mAP) on base categories. We evaluate the vanilla YOLO detector and our proposed detection model on three different sets of base categories.}
\label{tab:base}
\end{table}

\vspace{-3mm}
\subsection{Ablation Studies}


We analyze the effects of various components in our system, by comparing the performance on both base classes and novel classes. The experiments are on PASCAL VOC base/novel split 1, using 10-shot data on novel classes.

\minisection{Which layer output features to reweight.}   In our experiments, we apply the reweighting module to moderate the output of the second last layer (layer 21). This is the highest level of intermediate features we could use. However, other options could be considered as well. We experiment with applying the reweighting vectors to feature maps output from layer 20 and 13, while also considering only half of features in layer 21. The results are shown in Table \ref{tab:ablation_layer}. We can see that the it is more suitable to implement feature reweighting

at deeper layers, as using earlier layers gives  worse performance. Moreover, moderating only half of the features does not hurt the performance much, which demonstrates that a significant portion of features can be shared among classes, as we analyzed in Sec.~\ref{sec:vis}.

\begin{table}[!t]
    \small
    \centering
    \begin{tabular}{ccccc}
    \toprule
          & Layer 13 & Layer 20 & Layer 21 & Layer 21(half)\\ \midrule
    Base  & 69.6 & 69.2 & \textbf{69.7} & 69.2\\
    Novel & 40.7 & 43.6 & \textbf{47.2} & 46.9\\ 
    \bottomrule
    \end{tabular}
    \caption{Performance comparison for the detection models trained with reweighting applied on different layers.}
    \vspace{-3mm}
    \label{tab:ablation_layer}
\end{table}

\minisection{Loss functions.} As we mentioned in Sec.~\ref{sec:training_details},  there are several options for defining the classification loss. Among them the binary loss is the most straightforward one: if the inputs to the reweighting module and the detector are from the same class, the model predicts 1 and otherwise 0. This binary loss can be defined in following two ways. The single-binary loss refers to that in each iteration the reweighting module only takes one class of input, and the detector regresses 0 or 1; and the multi-binary loss refers to that per iteration the reweighting module takes $N$ examples from $N$ classes, and compute $N$ binary loss in total. Prior works on Siamese Network \cite{koch2015siamese} and Learnet \cite{bertinetto2016learning} use the single-binary loss. Instead, our model uses the softmax loss for calibrating the classification scores of $N$ classes. To investigate the effects of using different loss functions, we compare model performance trained with the single-binary, multi-binary loss and with our softmax loss in Table \ref{tab:ablation_loss}. We observe that using softmax loss significantly outperforms binary loss. This is likely due to its effect in suppressing redundant detection results.
\begin{table}[!t]
    \small
    \centering
    \begin{tabular}{cccc}
    \toprule
          & Single-binary & Multi-binary & Softmax\\ \midrule
    Base  & 49.1 & 64.1 & \textbf{69.7}\\
    Novel & 14.8 & 41.6 & \textbf{47.2} \\ 
    \bottomrule
    \end{tabular}
    \caption{Performance comparison for the detection models trained with different loss functions.}
    \label{tab:ablation_loss}
    \vspace{-3mm}
\end{table}



\minisection{Input form of reweighting module.} In our experiments, we use an image of the target class with a binary mask channel indicating   position of the object as input to the meta-model. We examine the case where we only feed the image. From Table \ref{tab:ablation_input} we see that this gives lower performance especially on novel classes. An apparently reasonable alternative is to feed the cropped target object together with the image. From Table \ref{tab:ablation_input}, this solution is also slightly worse. The necessity of the mask may lie in that it provides the precise information about the object location and its context.


We also analyze the  input sampling scheme for testing and   effect of sharing weights between feature extractor and reweighting module. See supplementary material.

\begin{table}[!h]
\small
    \definecolor{lightgray}{gray}{0.9}
    \centering
    \begin{tabular}{ccccccccc}
    \toprule
     Image & Mask & Object & Base & Novel \\ 
    \midrule
    \checkmark &  &   & 69.5 & 43.3 \\
    \rowcolor{lightgray}
     \checkmark & \checkmark &  & \textbf{69.7}  & \textbf{47.2} \\
     \checkmark &  & \checkmark & 69.2  & 45.8 \\
     \checkmark & \checkmark & \checkmark   & 69.4 & 46.8 \\ 
    \bottomrule
    \end{tabular}
        \caption{Performance comparison for different support input forms. The shadowed line is the one we use in  main experiments.}
    \label{tab:ablation_input}
    \vspace{-3mm}
\end{table}

\section{Conclusion}
This work is among the first to explore the practical and challenging few-shot detection problems. It introduced a new model to learn to fast adjust contributions  of the basic  features to detect novel classes with a few example. Experiments on realistic benchmark datasets clearly demonstrate its effectiveness. This work also compared the model learning speed, analyzed  predicted reweighting vectors and contributions of each design component,  providing  in-depth understanding of the proposed model. Few-shot detection is a challenging problem and we will further explore how to improve its performance for more complex scenes.

\section*{Acknolwedgement}
Jiashi Feng was partially supported by NUS IDS R-263-000-C67-646,  ECRA R-263-000-C87-133 and MOE Tier-II R-263-000-D17-112. This work was in part supported by the US DoD, the Berkeley Deep Drive  (BDD) Center, and the Berkeley Artificial Intelligence Research (BAIR) Lab.

{\small
\bibliographystyle{ieee_fullname}
\bibliography{egpaper_for_review}
}

\newpage
\section*{Implementation Details}
All our models are trained using SGD with momentum 0.9, and $L_2$ weight-decay 0.0005 (on both feature extractor and reweighting module). The batch size is set to be 64. For base training we train for 80,000 iterations, a step-wise learning rate decay strategy is used, with learning rate being 10$^{-4}$, 10$^{-3}$, 10$^{-4}$, 10$^{-5}$, and changes happening in iteration 500, 40,000, 60,000. For few-shot fine-tuning, we use a constant learning rate of 0.001 and train for 1500 iterations. We use multi-scale training, and evaluate the model in 416 $\times$ 416 resolution, as with the original YOLOv2.


\section*{Additional Ablation Studies}

\paragraph{Sampling of  Examples for Testing} 
During training, the reweighting module takes random input from the $k$-shot data each of the $N$ classes. In testing, we take the $k$-shot example as reweighting module's input and use the average of their predicted weights for detecting the corresponding class. If we replace the averaging process by randomly selecting reweighting module's input (as during training), the performance on base/novel classes will drop significantly from 69.7\%/47.2\% to 63.9\%/45.1\%. This is similar to the ensembling effect, except that this averaging over reweighting coefficients do not need additional inference time as in normal ensembling.

\vspace{-4mm}
\paragraph{Sharing Weights Between Feature Extractor and Reweighting Module} The first few layers of the reweighting module and the backbone feature extractor share the same architecture. Thus some weights can be shared between them. We evaluate this alternative and found the performance on base/novel classes decrease from 69.7\%/47.2\% to 68.3\%/44.8\%. The reason could be it imposes more constraints in the optimization process.

\section*{Complete Results on PASCAL VOC} Here we present the complete results for each class and number of shot on PASCAL VOC dataset. The results for base/novel split 1/2/3 are shown in Table 1/2/3 respectively.

\begin{table*}

\centering
\scriptsize
\setlength\tabcolsep{3pt}
\begin{tabular}{ll|cccccc|cccccccccccccccc}
\toprule
                    &             & \multicolumn{6}{c|}{Novel}               & \multicolumn{16}{c}{Base}                                                                                              \\ \hline
                 \# Shots     &           & bird & bus  & cow  & mbike & sofa & mean & aero & bike & boat & bottle & car  & cat  & chair & table & dog  & horse & person & plant & sheep & train & tv   & mean \\ \midrule
\multirow{6}{*}{1} 
					& \bs{}       & 0.0 & 0.0 & 0.0 & 0.0 & 0.0 & 0.0 & \textbf{78.4} & 76.9 & 61.5 & \textbf{48.7} & \textbf{79.8} & \textbf{84.5} & 51.0 & \textbf{72.7} & 79.0 & 77.6 & \textbf{74.9} & \textbf{48.2} & 62.8 & \textbf{84.8} & {73.1} & \textbf{70.2} \\ 
					& \bbsshort{} &6.8 & 0.0 & 9.1 & 0.0 & 0.0 & 3.2 & 77.1 & \textbf{78.2} & 61.7 & 46.7 & 79.4 & 82.7 & 51.0 & 69.0 & 78.3 & {79.5} & 74.2 & 42.7 & \textbf{68.3} & 84.1 & 72.9 & 69.7 \\ 
					& \bbs{} &11.4 & {17.6} & 3.8 & 0.0 & 0.0 & 6.6 & 75.8 & 77.3 & {63.1} & 45.9 & 78.7 & 84.1 & {52.3} & 66.5 & \textbf{79.3} & 77.2 & 73.7 & 44.0 & 66.0 & 84.2 & 72.2 & 69.4 \\
					& \lstd{} & 12.0 & 17.8 & 4.6 & 0.0 & 0.1 & 6.9 & 75.5 & 76.9 & \textbf{63.2} & 46.2 & 78.9 & 84.1 & \textbf{52.5} & 66.8 & 79.2 & 79.4 & 74.1 & 44.7 & 66.4 & 84.6 & \textbf{73.6} &	69.7 \\
					& \lstdfull{} & 13.4 & \textbf{21.4} & 6.3 & 0.0 & 0.0 & 8.2 & 73.4  & 73.5  & 61.8  & 44.7  & 78.4  & 83.9  & 50.8  & 68.3  & 79.3  & \textbf{80.5}  & 72.3  & 41.0  & 64.5  & 83.2  & 72.5  & 68.5 \\
					& \Ours  &\textbf{13.5} & 10.6 & \textbf{31.5} & \textbf{13.8} & \textbf{4.3} & \textbf{14.8} & 75.1 & 70.7 & 57.0 & 41.6 & 76.6 & 81.7 & 46.6 & 72.4 & 73.8 & 76.9 & 68.8 & 43.1 & 63.0 & 78.8 & 69.9 & 66.4 \\ \midrule
\multirow{6}{*}{2} 
					& \bs{}       & 0.0 & 0.0 & 0.0 & 0.0 & 0.0 & 0.0 & 77.6 & \textbf{77.6} & \textbf{60.4} & \textbf{48.1} & \textbf{81.5} & 82.6 & \textbf{51.5} & \textbf{72.0} & \textbf{79.2} & 78.8 & \textbf{75.2} & \textbf{47.0} & 65.2 & \textbf{86.0} & 72.7 & \textbf{70.4} \\ & \bbsshort{} &11.5 & 5.8 & 7.6 & 0.1 & 7.5 & 6.5 & \textbf{77.9} & 75.0 & 58.5 & 45.7 & 77.6 & {84.0} & 50.4 & 68.5 & 79.2 & {79.7} & 73.8 & 44.0 & \textbf{66.0} & 77.5 & \textbf{72.9} & 68.7 \\ & \bbs{} &16.6 & 9.7 & 12.4 & 0.1 & \textbf{14.5} & 10.7 & 76.4 & 70.2 & 56.9 & 43.3 & 77.5 & 83.8 & 47.8 & 70.7 & 79.1 & 77.6 & 71.7 & 39.6 & 61.4 & 77.0 & 70.3 & 66.9 \\ 
					& \lstd{} & 12.3  & 10.1  & 14.6  & 0.1  & 8.9  & 9.2 & 77.4  & 77.1  & 59.4  & 46.4  & 77.8  & \textbf{84.5}  & 50.9  & 67.1  & 79.1  & \textbf{80.6}  & 73.8  & 43.3  & 64.9  & 79.4  & 72.4  & 68.9 \\
					& \lstdfull{}  & 17.3  & 12.5  & 8.6  & 0.2  & 16.5  & 11.0  & 74.6  & 71.7  & 57.9  & 42.8  & 78.1  & 83.8  & 47.9  & 66.7  & 78.4  & 77.8  & 71.8  & 39.3  & 60.7  & 81.4  & 71.2  & 67.0 \\
					& \Ours  &\textbf{21.2} & \textbf{12.0} & \textbf{16.8} & \textbf{17.9} & 9.6 & \textbf{15.5} & 74.6 & 74.9 & 56.3 & 38.5 & 75.5 & 68.0 & 43.2 & 69.3 & 66.2 & 42.4 & 68.1 & 41.8 & 59.4 & 76.4 & 70.3 & 61.7 \\ \midrule

\multirow{6}{*}{3}  & \bs{}       & 0    & 0    & 0    & 0     & 9.1  & 1.8  & \textbf{78.0} & \textbf{77.2} & \textbf{61.2} & 45.6   & \textbf{81.6} & 83.7 & \textbf{51.7}  & \textbf{73.4}  & \textbf{80.7} & {79.6}  & \textbf{75.0}   & \textbf{45.5}  & 65.6  & 83.1  & {72.7} & \textbf{70.3} \\ 
                    & \bbsshort{} & 10.9 & 5.5  & 15.3 & 0.2   & 0.1  & 6.4  & 76.7 & 77.0 & 60.4 & \textbf{46.9}   & 78.8 & \textbf{84.9} & 51.0  & 68.3  & 79.6 & 78.7  & 73.1   & 44.5  & \textbf{67.6}  & \textbf{83.6}  & 72.4 & 69.6 \\ 
                    & \bbs{}      & 21.0 & 22.0 & 19.1 & 0.5   & 0.0  & 12.5 & 73.4 & 67.5 & 56.8 & 41.2   & 77.1 & 81.6 & 45.5  & 62.1  & 74.6 & 78.9  & 67.9   & 37.8  & 54.1  & 76.4  & 71.9 & 64.4 \\ 
                    & \lstd{} & 12.3  & 7.1  & 17.7  & 0.1  & 0.0  & 7.5  & 75.9  & 76.2  & 59.7  & 46.6  & 78.3  & 84.4  & 49.4  & 64.5  & 78.7  & \textbf{79.7}  & 72.6  & 42.5  & 63.8  & 80.5  & \textbf{73.}9  & 68.4 \\
                    & \lstdfull{}  & 23.1  & \textbf{22.6}  & 15.9  & 0.4  & 0.0  & 12.4  & 74.8  & 68.7  & 57.1  & 44.1  & 78.0  & 83.4  & 46.9  & 64.0  & 78.7  & 79.1  & 70.1  & 39.2  & 58.1  & 79.8  & 71.9  & 66.3 \\
                    & \Ours{}     & \textbf{26.1} & {19.1} & \textbf{40.7} & \textbf{20.4}  & \textbf{27.1} & \textbf{26.7} & 73.6 & 73.1 & 56.7 & 41.6   & 76.1 & 78.7 & 42.6  & 66.8  & 72.0 & 77.7  & 68.5   & 42.0  & 57.1  & 74.7  & 70.7 & 64.8 \\ \midrule
\multirow{4}{*}{5} 
					& \bs{}       & 0.0 & 0.0 & 0.0 & 0.0 & 9.1 & 1.8 & \textbf{77.8} & 76.4 & \textbf{65.7} & \textbf{45.9} & \textbf{79.5} & 82.3 & {50.4} & \textbf{72.5} & \textbf{79.1} & {79.0} & \textbf{75.5} & \textbf{47.9} & \textbf{67.2} & 83.0 & \textbf{72.5} & \textbf{70.3} \\ & \bbsshort{} &11.6 & 7.1 & 10.7 & 2.1 & 6.0 & 7.5 & 76.5 & \textbf{76.4} & 61.0 & 45.5 & 78.7 & {84.5} & 49.2 & 68.7 & 78.5 & 78.1 & 73.7 & 45.4 & 66.8 & \textbf{85.3} & 70.0 & 69.2 \\ & \bbs{} &20.2 & 20.0 & 22.4 & 36.4 & 24.8 & 24.8 & 72.0 & 70.6 & 60.7 & 42.0 & 76.8 & 84.2 & 47.7 & 63.7 & 76.9 & 78.8 & 72.1 & 42.2 & 61.1 & 80.8 & 69.9 & 66.6 \\ 
					& \lstd{} & 12.9  & 8.1  & 13.6  & 16.1  & 10.2  & 12.2  & 77.4  & 75.0  & 61.1  & 45.2  & 78.4  & \textbf{85.0}  & \textbf{50.6}  & 68.0  & 78.1  & \textbf{79.3}  & 73.1  & 44.6  & 65.5  & 84.5  & 71.1  & 69.1 \\
					& \lstdfull{} & 24.1  & \textbf{30.2}  & 24.0  & 40.0  & 25.6  & 29.1  & 74.2  & 70.7  & 60.4  & 42.9  & 77.3  & 83.1  & 47.9  & 66.0  & 76.9  & 79.2  & 71.3  & 41.4  & 61.0  & 80.2  & 70.2  & 66.8 \\
					& \Ours  &\textbf{31.5} & {21.1} & \textbf{39.8} & \textbf{40.0} & \textbf{37.0} & \textbf{33.9} & 69.3 & 57.5 & 56.8 & 37.8 & 74.8 & 82.8 & 41.2 & 67.3 & 74.0 & 77.4 & 70.9 & 40.9 & 57.3 & 73.5 & 69.3 & 63.4 \\ \midrule                    
\multirow{6}{*}{10} & \bs{}       & 0.0  & 0.0  & 0.0  & 0.0   & 9.1  & 1.8  & 76.9 & \textbf{77.1} & \textbf{62.2} & \textbf{47.3}   & \textbf{79.4} & \textbf{85.1} & \textbf{51.3}  & \textbf{70.1}  & {78.6} & 78.0  & \textbf{75.2}   & \textbf{47.4}  & 63.9  & \textbf{85.0}  & \textbf{72.3} & \textbf{70.0} \\ 
                    & \bbsshort{} & 11.4 & 28.4 & 8.9  & 4.8   & 7.8  & 12.2 & \textbf{77.4} & 76.9 & 60.9 & 44.8   & 78.3 & 83.2 & 48.5  & 68.9  & 78.5 & 78.9  & 72.6   & 44.8  & \textbf{67.3}  & 82.7  & 69.3 & 68.9 \\ 
                    & \bbs{}      & 22.3 & 53.9 & 32.9 & 40.8  & 43.2 & 38.6 & 71.9 & 69.8 & 57.1 & 41.0   & 76.9 & 81.7 & 43.6  & 65.3  & 77.3 & \textbf{79.2}  & 70.1   & 41.5  & 63.7  & 76.9  & 69.1 & 65.7 \\ 
                    & \lstd{} & 11.3  & 32.2  & 5.6  & 1.3  & 7.7  & 11.6  & 77.1  & 75.2  & 62.0  & 44.5  & 78.2  & 84.2  & 49.9  & 68.6  & \textbf{78.8}  & 78.8  & 72.6  & 45.0  & 66.9  & 82.6  & 69.5  & 68.9 \\
                    & \lstdfull{} & 22.8  & 52.5  & 31.3  & 45.6  & 40.3  & 38.5  & 70.9  & 71.3  & 59.8  & 41.1  & 77.1  & 81.9  & 45.1  & 67.2  & 78.0  & 78.9  & 70.7  & 41.6  & 63.8  & 79.7  & 66.8  & 66.3 \\
                    & \Ours       & \textbf{30.0} & \textbf{62.7} & \textbf{43.2} & \textbf{60.6}  & \textbf{40.6} & \textbf{47.2} & 65.3 & 73.5 & 54.7 & 39.5   & 75.7 & 81.1 & 35.3  & 62.5  & 72.8 & 78.8  & 68.6   & 41.5  & 59.2  & 76.2  & 69.2 & 63.6 \\ \midrule

\end{tabular}
\caption{Detection performance (AP) for the base and novel categories on the PASCAL VOC dataset for the 1st base/novel split. We evaluate the performance for different numbers of training examples for the novel categories. }
\vspace{-3mm}
\label{tab:detail1}
\end{table*}

\begin{table*}

\centering
\scriptsize
\setlength\tabcolsep{3pt}
\begin{tabular}{ll|cccccc|cccccccccccccccc}
\toprule
                    &             & \multicolumn{6}{c|}{Novel}               & \multicolumn{16}{c}{Base}                                                                                              \\ \hline
                 \# Shots     &           & aero & bottle & cow & horse & sofa & mean & bike & bird & boat & bus & car & cat & chair & table & dog & mbike & person & plant & sheep & train & tv   & mean \\ \midrule
\multirow{6}{*}{1} 
					& \bs{}       & 0.0 & 0.0 & 0.0 & 0.0 & 0.0 & 0.0 & \textbf{78.8} & \textbf{73.2} & \textbf{63.6} & 79.0 & {79.7} & \textbf{87.2} & 51.5 & 71.2 & \textbf{81.1} & 78.1 & \textbf{75.4} & \textbf{47.7} & 65.9 & \textbf{84.0} & 73.7 & {72.7} \\ & \bbsshort{} &0.4 & 0.2 & 10.3 & 29.8 & 0.0 & 8.2 & 77.9 & 70.2 & 62.2 & \textbf{79.8} & 79.4 & 86.6 & \textbf{51.9} & 72.3 & 77.1 & 78.1 & 73.9 & 44.1 & \textbf{66.6} & 83.4 & {74.0} & 71.8 \\ 
					& \bbs{} &0.6 & \textbf{9.1} & 11.2 & {41.6} & 0.0 & 12.5 & 74.9 & 67.2 & 60.1 & 78.8 & 79.0 & 83.8 & 50.6 & \textbf{72.7} & 75.5 & 74.8 & 71.7 & 43.9 & 62.5 & 81.8 & 72.6 & 70.0 \\ 
					& \lstd{} & 0.5  & 0.1  & 11.1  & 37.7  & 0.0  & 9.9  & 76.9  & 69.8  & 61.5  & 78.2  & \textbf{81.0}  & 85.7  & 51.9  & \textbf{73.7}  & 79.6  & 76.7  & 73.4  & 43.8  & 66.0  & 82.2  & \textbf{74.1}  & 71.6 \\
					& \lstdfull{} & 0.1  & 1.5  & 10.4  & \textbf{44.9}  & 0.0  & 11.4  & 76.1  & 68.0  & 58.7  & 78.1  & 79.0  & 85.0  & 50.7  & 72.2  & 76.2  & 75.2  & 71.8  & 43.3  & 62.7  & 82.8  & 72.2  & 70.1 \\
					& \Ours  &\textbf{11.8} & \textbf{9.1} & \textbf{15.6} & 23.7 & \textbf{18.2} & \textbf{15.7} & 77.6 & 62.7 & 54.2 & 75.3 & 79.0 & 80.0 & 49.6 & 70.3 & 78.3 & \textbf{78.2} & 68.5 & 42.2 & 58.2 & 78.5 & 70.4 & 68.2 \\ \midrule

\multirow{6}{*}{2} 
					& \bs{}       & 0.0 & 0.6 & 0.0 & 0.0 & 0.0 & 0.1 & \textbf{78.4} & 69.7 & \textbf{64.5} & 78.3 & \textbf{79.7} & 86.1 & \textbf{52.2} & \textbf{72.6} & \textbf{81.2} & 78.6 & \textbf{75.2} & \textbf{50.3} & \textbf{66.1} & \textbf{85.3} & 74.0 & \textbf{72.8} \\ 
					& \bbsshort{} &0.2 & 0.2 & 17.2 & 1.2 & 0.0 & 3.8 & 78.1 & {70.0} & 60.6 & 79.8 & 79.4 & \textbf{87.1} & 49.7 & 70.3 & 80.4 & \textbf{78.8} & 73.7 & 44.2 & 62.2 & 82.4 & {74.9} & 71.4 \\ 
					& \bbs{} &1.8 & {1.8} & 15.5 & \textbf{1.9} & 0.0 & 4.2 & 76.4 & 69.7 & 58.0 & \textbf{80.0} & 79.0 & 86.9 & 44.8 & 68.2 & 75.2 & 77.4 & 72.2 & 40.3 & 59.1 & 81.6 & 73.4 & 69.5 \\ 
					& \lstd{} & 0.4  & \textbf{4.5}  & 21.5  & 0.5  & 0.0  & 5.4  & 77.5  & \textbf{71.8}  & 61.4  & 79.5  & 79.4  & 86.9  & 48.6  & 71.0  & 80.1  & 77.2  & 74.0  & 43.3  & 63.6  & 81.8  & \textbf{75.3}  & 71.4 \\
					& \lstdfull{} & 3.0  & 1.5  & 13.9  & 0.6  & 0.0  & 3.8  & 77.2  & 69.0  & 58.2  & 77.6  & 79.1  & 86.3  & 45.6  & 70.2  & 77.1  & 76.3  & 72.7  & 40.3  & 59.4  & 81.1  & 74.4  & 69.6 \\
					& \Ours  &\textbf{28.6} & 0.9 & \textbf{27.6} & 0.0 & \textbf{19.5} & \textbf{15.3} & 75.8 & 67.4 & 52.4 & 74.8 & 76.6 & 82.5 & 44.5 & 66.0 & 79.4 & 76.2 & 68.2 & 42.3 & 53.8 & 76.6 & 71.0 & 67.2 \\ \midrule
\multirow{4}{*}{3} 
					& \bs{}       & 0.0 & 0.0 & 0.0 & 0.0 & 0.0 & 0.0 & 77.6 & \textbf{72.2} & 61.2 & 77.9 & 79.8 & 85.8 & \textbf{49.9} & \textbf{73.2} & \textbf{80.0} & 77.9 & \textbf{75.3} & \textbf{50.8} & \textbf{64.3} & \textbf{84.2} & 72.6 & \textbf{72.2} \\ & \bbsshort{} &4.9 & 0.0 & 11.2 & 1.2 & 0.0 & 3.5 & \textbf{78.7} & 71.6 & \textbf{62.4} & 77.4 & \textbf{80.4} & {87.5} & 49.5 & 70.8 & 79.7 & \textbf{79.5} & 72.6 & 44.3 & 60.0 & 83.0 & {75.2} & 71.5 \\
					& \bbs{} &10.7 & \textbf{4.6} & 12.9 & \textbf{29.7} & 0.0 & 11.6 & 74.9 & 69.2 & 60.4 & {79.4} & 79.1 & 87.3 & 43.4 & 69.7 & 75.8 & 75.2 & 70.5 & 39.4 & 52.9 & 80.8 & 73.4 & 68.8 \\
					& \lstd{} & 4.5  & 0.1  & 10.8  & 0.8  & 0.0  & 3.2  & 78.4  & 71.5  & 60.9  & 78.5  & 80.2  & 87.7  & 47.8  & 70.4  & {79.5}  & 77.8  & 73.1  & 42.9  & 58.9  & 81.6  & \textbf{75.4}  & 71.0 \\
					& \lstdfull{}  & 12.6  & 0.7  & 11.3  & 0.4  & 0.0  & 5.0  & 75.5  & 69.7  & 61.0  & \textbf{79.5}  & 79.1  & \textbf{87.8}  & 43.2  & 68.5  & 76.0  & 75.7  & 71.0  & 41.2  & 61.2  & 80.9  & 73.3  & 69.6 \\
					& \Ours  &\textbf{29.4} & \textbf{4.6} & \textbf{34.9} & 6.8 & \textbf{37.9} & \textbf{22.7} & 62.6 & 64.7 & 55.2 & 76.6 & 77.1 & 82.7 & 46.7 & 65.4 & 75.4 & 78.3 & 69.2 & 42.8 & 45.2 & 77.9 & 69.6 & 66.0 \\ \midrule

\multirow{6}{*}{5} 
					& \bs{}       & 0.0 & 0.0 & 0.0 & 0.0 & 9.1 & 1.8 & 78.0 & 71.5 & \textbf{62.9} & \textbf{81.7} & 79.7 & 86.8 & \textbf{50.0} & \textbf{72.3} & \textbf{81.7} & 77.9 & \textbf{75.6} & \textbf{48.4} & \textbf{65.4} & 83.2 & 73.6 & \textbf{72.6} \\ & \bbsshort{} &0.8 & 0.2 & 11.3 & 5.2 & 0.0 & 3.5 & {78.6} & \textbf{72.4} & 61.5 & 79.4 & \textbf{81.0} & \textbf{87.8} & 48.6 & 72.1 & 81.0 & \textbf{79.6} & 73.6 & 44.9 & 61.4 & \textbf{83.9} & \textbf{74.7} & 72.0 \\ & \bbs{} &10.3 & 9.1 & 17.4 & \textbf{43.5} & 0.0 & 16.0 & 76.4 & 69.6 & 59.1 & 80.3 & 78.5 & 87.8 & 42.1 & 72.1 & 76.6 & 77.1 & 70.7 & 43.1 & 58.0 & 82.4 & 72.6 & 69.8 \\ 
					& \lstd{}  & 0.7  & 0.6  & 13.0  & 14.3  & 0.0  & 5.7  & \textbf{79.1}  & 72.4  & 62.0  & 78.6  & 80.8  & 87.2  & 44.9  & 71.3  & 79.3  & 78.3  & 72.4  & 44.5  & 62.1  & 82.1  & 74.7  & 71.3 \\
					& \lstdfull{} & 11.6  & 9.1  & 15.2  & 42.9  & 0.0  & 15.8  & 76.4  & 70.7  & 59.4  & 77.5  & 78.9  & 87.6  & 41.6  & 70.7  & 76.8  & 77.8  & 70.2  & 42.1  & 57.9  & 82.8  & 72.3  & 69.5 \\
					& \Ours  &\textbf{33.1} & \textbf{9.4} & \textbf{38.4} & 25.4 & \textbf{44.0} & \textbf{30.1} & 73.2 & 65.6 & 52.9 & 75.9 & 77.5 & 80.0 & 43.7 & 65.0 & 73.8 & 78.4 & 68.9 & 39.2 & 56.4 & 78.0 & 70.8 & 66.6 \\ \midrule
\multirow{6}{*}{10} 
					& \bs{}       & 0.0 & 0.0 & 0.0 & 0.0 & 0.0 & 0.0 & 77.4 & 71.5 & 61.1 & \textbf{78.8} & \textbf{82.7} & \textbf{87.1} & \textbf{52.5} & \textbf{74.6} & \textbf{80.8} & \textbf{79.3} & \textbf{75.4} & \textbf{46.1} & 64.2 & \textbf{85.2} & 73.6 & \textbf{72.7} \\ & \bbsshort{} &3.8 & 0.0 & 18.3 & 17.0 & 0.0 & 7.8 & \textbf{79.3} & \textbf{72.8} & {61.6} & 78.5 & 81.4 & 87.1 & 46.9 & 73.3 & 79.8 & 79.0 & 73.1 & 44.6 & {65.9} & 83.4 & \textbf{73.7} & 72.0 \\ & \bbs{} &41.7 & 9.5 & 34.5 & 45.1 & 38.4 & 33.9 & 75.5 & 69.4 & 60.0 & 78.3 & 78.8 & 86.8 & 44.9 & 68.4 & 75.8 & 76.9 & 70.7 & 44.0 & 64.1 & 81.6 & 71.1 & 69.8 \\
					& \lstd{} & 31.2  & 9.1  & 22.3  & 25.6  & 7.8  & 19.2  & 78.8  & 72.5  & \textbf{62.3}  & 78.5  & 80.9  & 86.8  & 47.4  & 70.8  & 79.6  & 78.6  & 72.7  & 44.2  & \textbf{66.5}  & 83.7  & 73.3  & 71.8 \\
					& \lstdfull{} & 41.5  & 9.3  & 29.2  & 38.9  & 36.1  & 31.0  & 74.6  & 70.2  & 59.6  & 77.3  & 78.6  & 86.5  & 45.1  & 68.1  & 77.6  & 75.2  & 70.6  & 44.5  & 59.8  & 79.7  & 71.2  & 69.2 \\
					& \Ours  & \textbf{41.8}  & \textbf{14.0}  & \textbf{42.7}  & \textbf{63.4 } & \textbf{40.7}  & \textbf{40.5}  & 75.2  & 65.2  & 46.7  & 74.9  & 78.5  & 79.1  & 36.0  & 58.4  & 73.0  & 77.7  & 67.9  & 39.9  & 57.1  & 75.2  & 66.3  & 64.7 \\ \midrule

\end{tabular}
\caption{Detection performance (AP) for the base and novel categories on the PASCAL VOC dataset for the 2nd base/novel split. We evaluate the performance for different numbers of training examples for the novel categories. }
\vspace{-3mm}
\label{tab:detail2}
\end{table*}

\begin{table*}[!t]
\centering
\scriptsize
\setlength\tabcolsep{3pt}
\begin{tabular}{ll|cccccc|cccccccccccccccc}
\toprule
                    &             & \multicolumn{6}{c|}{Novel}               & \multicolumn{16}{c}{Base}                                                                                              \\ \hline
                 \# Shots     &           & boat & cat & mbike & sheep & sofa & mean & aero & bike & bird & bottle & bus & car & chair & cow & table & dog & horse & person & plant & train & tv & mean \\ \midrule
\multirow{6}{*}{1} 
					& \bs{}       & 0.0 & 9.1 & 0.0 & 0.0 & 0.0 & 1.8 & \textbf{78.7} & \textbf{76.8} & \textbf{73.4} & \textbf{48.8} & \textbf{79.0} & \textbf{82.3} & \textbf{50.2} & 68.4 & \textbf{71.4} & \textbf{76.7} & \textbf{80.7} & \textbf{75.0} & \textbf{46.8} & \textbf{83.8} & \textbf{71.7} & \textbf{70.9} \\ 
					& \bbsshort{} &0.1 & 25.8 & 10.7 & 3.6 & 0.1 & 8.1 & 77.2 & 74.9 & 69.1 & 47.4 & 78.7 & 79.7 & 47.9 & 68.3 & 69.6 & 74.7 & 79.4 & 74.2 & 42.2 & 82.7 & 71.1 & 69.1 \\ 
					& \bbs{} &0.1 & 30.9 & {26.0} & 8.0 & 0.1 & 13.0 & 75.1 & 70.7 & 65.9 & 43.6 & 78.4 & 79.5 & 47.8 & \textbf{68.7} & 68.0 & 72.8 & 79.5 & 72.3 & 40.1 & 80.5 & 68.6 & 67.4 \\
					& \lstd & 0.1  & 30.8  & 17.5  & 6.0  & 0.1  & 10.9  & 76.3  & 74.8  & 68.2  & 45.6  & 77.2  & 80.0  & 48.6  & 70.1  & 69.0  & 71.5  & 79.9  & 73.7  & 42.0  & 81.3  & 70.1  & 68.5 \\
					& \lstdfull & 0.0  & 27.8  & 25.0  & 9.7  & 0.2  & 12.6  & 75.8  & 71.7  & 65.1  & 44.0  & 78.1  & 79.3  & 46.7  & 68.0  & 68.9  & 68.1  & 79.0  & 72.4  & 40.2  & 80.2  & 68.3  & 67.1 \\
					& \Ours & \textbf{10.3}  & \textbf{41.4}  & \textbf{29.1}  & \textbf{16.2}  & \textbf{9.4} & \textbf{21.3}  & 77.6  & 72.6  & 65.7  & 39.6  & 77.0  & 78.2  & 49.7  & 53.9  & 64.6  & 67.4  & 79.3  & 67.2  & 41.0  & 82.5  & 72.5  & 65.9 \\ \midrule
\multirow{6}{*}{2} 
					& \bs{}       & 0.0 & 9.1 & 0.0 & 0.0 & 0.0 & 1.8 & 77.6 & \textbf{77.1} & \textbf{74.0} & \textbf{49.4} & \textbf{79.8} & \textbf{79.9} & \textbf{50.5} & 71.0 & \textbf{72.7} & \textbf{76.3} & 81.0 & \textbf{75.0} & \textbf{48.4} & \textbf{84.9} & \textbf{72.7} & \textbf{71.4} \\ & \bbsshort{} &0.0 & 24.4 & 2.5 & 9.8 & 0.1 & 7.4 & \textbf{78.2} & 76.0 & 72.2 & 47.2 & 79.3 & 79.8 & 47.3 & \textbf{72.1} & 70.0 & 74.9 & 80.3 & 74.3 & 45.2 & 84.9 & 72.0 & 70.2 \\ & \bbs{} &0.0 & 35.2 & \textbf{28.7} & 15.4 & 0.1 & 15.9 & 75.3 & 72.0 & 69.8 & 44.0 & 79.1 & 78.8 & 42.1 & 70.0 & 64.9 & 73.8 & {81.7} & 71.4 & 40.9 & 80.9 & 69.4 & 67.6 \\ 
					& \lstd & 0.0  & 25.4  & 0.0  & 12.6  & 0.1  & 7.6  & 78.1  & 76.4  & 71.9  & 46.8  & 78.8  & 79.6  & 45.3  & 70.6  & 66.9  & 75.3  & 81.7  & 73.9  & 43.4  & 84.1  & 71.8  & 69.7 \\
					& \lstdfull & 0.2  & 27.3  & 0.1  & 15.0  & 0.2  & 8.5  & 77.4  & 73.3  & 69.5  & 44.8  & 78.5  & 79.2  & 43.0  & 69.2  & 66.4  & 71.9  & \textbf{82.0}  & 72.3  & 39.8  & 84.5  & 69.3  & 68.1 \\
					& \Ours & \textbf{6.3}  & \textbf{47.1}  & 28.4  & \textbf{28.1}  & \textbf{18.2}  & \textbf{25.6}  & 75.8  & 73.0  & 66.4  & 40.0  & 77.8  & 77.6  & 43.1  & 62.6  & 58.5  & 71.0  & 78.9  & 67.0  & 41.2  & 77.0  & 70.0  & 65.3 \\ \midrule
\multirow{6}{*}{3} 
					& \bs{}       & 0.0 & 9.1 & 0.0 & 0.0 & 0.0 & 1.8 & 77.1 & \textbf{77.0} & 70.6 & 46.3 & 77.5 & 79.7 & \textbf{49.7} & 68.8 & \textbf{73.4} & 74.5 & 79.4 & \textbf{75.6} & \textbf{48.1} & {83.6} & \textbf{72.1} & \textbf{70.2} \\
					& \bbsshort{} &0.0 & 27.0 & 1.8 & 9.1 & 0.1 & 7.6 & \textbf{77.7} & 76.6 & \textbf{71.4} & {47.5} & 78.0 & \textbf{79.9} & 47.6 & \textbf{70.0} & 70.5 & 74.4 & \textbf{80.0} & 73.7 & 44.1 & 83.0 & 70.9 & 69.7 \\ &
					\bbs{} &0.0 & 39.0 & 18.1 & 17.9 & 0.0 & 15.0 & 73.2 & 71.1 & 68.8 & 43.7 & \textbf{78.9} & 79.3 & 43.1 & 67.8 & 62.2 & \textbf{76.3} & 79.4 & 70.8 & 40.5 & 81.6 & 69.6 & 67.1 \\
					& \lstd & 0.0  & 29.0  & 9.5  & 9.1  & 0.1  & 9.5  & 77.7  & 76.2  & 69.8  & \textbf{48.1}  & 77.9  & 79.9  & 46.9  & 69.7  & 69.0  & 75.0  & 79.9  & 73.8  & 43.9  & \textbf{83.8}  & 70.9  & 69.5 \\
					& \lstdfull  & 0.0  & 36.6  & 21.4  & 16.9  & 0.0  & 15.0  & 74.7  & 73.2  & 67.7  & 44.7  & 78.1  & 79.5  & 40.5  & 69.0  & 60.9  & 76.0  & 79.1  & 71.0  & 40.1  & 83.0  & 69.6  & 67.1 \\
					& \Ours  & \textbf{11.7}  & \textbf{48.2}  & \textbf{17.4}  & \textbf{34.7}  & \textbf{30.1}  & \textbf{28.4}  & 73.2  & 69.3  & 66.5  & 41.8  & 77.6  & 76.3  & 42.8  & 61.1  & 63.7  & 67.3  & 77.4  & 68.2  & 39.7  & 78.6  & 70.7  & 65.0 \\ \midrule

\multirow{6}{*}{5} 
					& \bs{}       & 0.0 & 9.1 & 0.0 & 0.0 & 9.1 & 3.6 & \textbf{78.2} & \textbf{78.5} & \textbf{72.1} & \textbf{47.8} & 76.6 & \textbf{82.1} & \textbf{50.7} & 70.1 & \textbf{71.8} & {77.6} & \textbf{80.4} & \textbf{75.4} & \textbf{46.0} & \textbf{84.8} & \textbf{72.5} & \textbf{71.0} \\ & \bbsshort{} &0.0 & 33.8 & 2.6 & 7.8 & 3.2 & 9.5 & 77.2 & 77.1 & 71.9 & 47.3 & \textbf{78.8} & 79.8 & 47.1 & 69.8 & 71.8 & 77.0 & 80.2 & 74.3 & 44.2 & 82.5 & 70.6 & 70.0 \\ & \bbs{} &7.9 & 48.0 & 39.1 & 29.4 & 36.6 & 32.2 & 75.5 & 73.6 & 69.1 & 43.3 & 78.4 & 78.9 & 42.3 & \textbf{70.2} & 66.1 & 77.4 & 79.8 & 72.2 & 41.9 & 82.8 & 69.3 & 68.1 \\ 
					& \lstd & 0.0  & 39.1  & 12.4  & 15.8  & 9.2  & 15.3  & 77.6  & 76.8  & 71.0  & 46.3  & 78.2  & 79.9  & 46.2  & 71.3  & 69.8  & \textbf{77.7}  & 80.0  & 74.3  & 45.1  & 83.1  & 71.2  & 69.9 \\
					& \lstdfull & 0.2  & 51.5  & 37.2  & 26.9  & 20.7  & 27.3  & 74.5  & 73.5  & 69.1  & 42.9  & 78.4  & 79.2  & 42.3  & 69.3  & 66.0  & \textbf{77.7}  & 79.6  & 71.7  & 41.6  & 82.8  & 69.2  & 67.8 \\
					& \Ours & \textbf{14.8}  & \textbf{59.1}  & \textbf{49.6 } & \textbf{45.0 } &\textbf{ 45.6}  & \textbf{42.8}  & 70.4  & 69.3  & 65.9  & 40.7  & 76.6  & 77.4  & 43.0  & 63.5  & 63.8  & 68.9  & 79.6  & 71.5  & 44.1  & 80.9  & 69.7  & 65.7 \\ \midrule
					
\multirow{6}{*}{10} 
					& \bs{}       & 0.0 & 9.1 & 1.5 & 0.0 & 9.1 & 3.9 & \textbf{78.7} & 77.1 & \textbf{73.3} & \textbf{48.0} & \textbf{79.4} & \textbf{79.8} & \textbf{51.6} & \textbf{71.7} & \textbf{71.1} & {77.6} & 79.9 & \textbf{74.4} & \textbf{47.8} & 83.2 & \textbf{73.4} & \textbf{71.1} \\ 
					& \bbsshort{} &0.0 & 35.6 & 1.0 & 14.2 & 1.5 & 10.5 & 78.0 & \textbf{77.8} & 69.0 & 46.0 & 78.5 & 79.6 & 45.3 & 69.9 & 70.9 & 77.0 & 80.8 & 74.2 & 45.2 & \textbf{83.3} & 70.7 & 69.8 \\ & 
					\bbs{} &12.0 & {59.6} & 42.5 & 39.1 & {38.9} & 38.4 & 73.2 & 74.0 & 66.5 & 44.0 & 78.1 & 78.5 & 43.6 & 68.0 & 66.9 & 76.9 & {81.4} & 72.1 & 43.8 & 82.1 & 68.2 & 67.8 \\ 
					& \lstd  & 0.0  & 37.0  & 18.6  & 22.7  & 6.5  & 16.9  & 77.2  & 77.2  & 68.5  & 44.4  & 78.7  & 79.3  & 44.4  & 69.9  & 70.2  & \textbf{78.6}  & 80.1  & 73.3  & 43.6  & 81.4  & 69.7  & 69.1 \\
					& \lstdfull  & 10.5  & 53.3  & 41.9  & 36.2  & 39.5  & 36.3  & 75.5  & 74.0  & 66.3  & 43.6  & 77.6  & 78.9  & 41.9  & 65.8  & 66.5  & 77.5  & \textbf{81.7}  & 71.7  & 42.5  & 80.6  & 67.5  & 67.4 \\
					& \Ours & \textbf{22.4}  & \textbf{59.9 } & \textbf{59.8}  & \textbf{46.1}  & \textbf{41.4}  & \textbf{45.9}  & 69.4  & 67.6  & 67.1  & 39.3  & 69.5  & 76.6  & 33.4  & 60.1  & 58.5  & 70.1  & 79.2  & 67.8  & 43.5  & 78.8  & 65.0  & 63.1 \\ \midrule

\end{tabular}
\caption{Detection performance (AP) for the base and novel categories on the PASCAL VOC dataset for the 3rd base/novel split. We evaluate the performance for different numbers of training examples for the novel categories. }
\vspace{-3mm}
\label{tab:detail3}
\end{table*}

\end{document}